\documentclass[11pt,a4paper]{article}
\usepackage{enumitem}
\usepackage[hyperref]{acl2020}
\usepackage{times}
\usepackage{latexsym}
\usepackage{textcomp}
\usepackage[english=american]{csquotes} 

\usepackage{multirow} 
\usepackage{float}
\usepackage{booktabs}
\usepackage{tikz}
\usetikzlibrary{positioning,chains}
\usepackage{multirow} 
\usepackage{amsmath,amsfonts,amssymb,amsthm}
\usepackage[figuresright]{rotating}
\usepackage{float}
\usepackage{rotating}
\setquotestyle{english}
\SetBlockThreshold{1} 
\usepackage{fontawesome}
\usepackage{caption}
\usepackage{subcaption}
\usepackage{pgfplots}
\usepgfplotslibrary{statistics}
\usepgfplotslibrary{colorbrewer}
\pgfplotsset{
    cycle list/Dark2,
    cycle multiindex* list={
        mark list*\nextlist
        Dark2\nextlist
    },
}
\usepackage[nohyperlinks,printonlyused]{acronym}
\makeatletter
\def\blfootnote{\gdef\@thefnmark{}\@footnotetext}
\makeatother
\usepackage{tcolorbox}
\tcbuselibrary{skins}
\makeatletter
\def\blfootnote{\gdef\@thefnmark{}\@footnotetext}
\makeatother
\newcommand{\tweet}[1] {
    \begin{tcolorbox}[
                    skin=enhanced,
                    width=3.0in,
                    colback=white,
                    fontlower=\sffamily,
                    fontupper=~\rmfamily,
                    middle=0mm,
                    center,
                    bottom=0.1pt
                    ]
    \includegraphics[width=0.5cm]{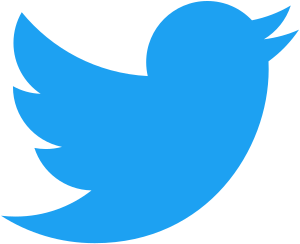}
    #1
    \end{tcolorbox}
}

\definecolor{orangeX}{rgb}{1,.5,0}
\definecolor{blueX}{rgb}{.2, .59, .88}
\definecolor{purpleX}{rgb}{.294118, 0, .509804}
\definecolor{greenX}{rgb}{.721, .878, .341}
\definecolor{bole}{rgb}{0.47, 0.27, 0.23}
\definecolor{greenish}{rgb}{0.0,0.7,0.0}
\usepackage{xspace}

\newcommand{\eg}{e.\,g.\,, }

\newcommand{\cf}{{cf.\,}}
\newcommand{\axel}{\texttt{AXEL}\xspace}
\newcommand{\wip}{\texttt{XLM1L\_AXEL}\xspace}

\newcommand{\bert}{\texttt{BERT}\xspace}
\newcommand{\xlm }{\texttt{XLM}\xspace}
\newcommand{\fast}{fastText\xspace}
\newcommand{\fasta}{fastText aligned\xspace}
\newcommand{\glo}{GloVe\xspace}
\newcommand{\hs}{hate speech\xspace}
\newcommand{\hsd}{hate speech detection\xspace}
\usepackage{soul}

\usepackage{microtype}

\aclfinalcopy 

\title{Cross-lingual Zero- and Few-shot Hate Speech Detection \linebreak utilising frozen Transformer Language Models and \axel}

\author{Lukas Stappen\thanks{   equal contribution} \\
  University of Augsburg, GER\\
  \texttt{stappen@ieee.org} \\ \And
  Fabian Brunn\footnotemark[1] \\
  TU Munich, GER \\
  \texttt{fabian.brunn@tum.de} \\ \And
  Bj{\"o}rn Schuller \\
  Imperial College London, UK\\
  \texttt{schuller@ieee.org} \\}

\date{}
\pgfplotsset{compat=1.13} 
\begin{document}
\maketitle
\begin{abstract}
Detecting hate speech, especially in low-resource languages, is a non-trivial challenge.
To tackle this, we developed a tailored architecture based on frozen, pre-trained Transformers to examine cross-lingual zero-shot and few-shot learning, in addition to uni-lingual learning, on the HatEval challenge data set. With our novel attention-based classification block \axel, we demonstrate highly competitive results on the English and Spanish subsets. We also re-sample the English subset, enabling additional, meaningful comparisons in the future.
\end{abstract}

\section{Introduction}
Hate speech, discriminatory communication intended to insult and intimidate specific groups or individuals due to their gender, race, sexual orientation or other characteristics has been a negative side effect of the growth of social media. Its effects are not confined to the virtual world; offline, it can result in criminal acts, including physical attacks \citep{muller2018fanning,Laub2019}.  In an extreme example, it has been heavily implicated in inciting violence against Rohingya Muslims in Myanmar in 2017 \citep{stevenson_2018, bbc_news_2018, reports_2018}, which included the murder of thousands of civilians and created close to a million refugees \citep{HRC2018}.

With billions of text snippets posted daily on social media, detecting hate speech using human observers is unfeasible, motivating researchers to develop natural language processing (NLP) methods to automate the task. Attempts by industry have so far fallen short; according to Facebook, their detection algorithms failed in the lead-up to the Rohingya crisis due to a lack of training data in Burmese \citep{murphy_2019}.

 
Detection approaches that work effectively with small or non-existent training data sets in the target language, such as cross-lingual zero- and few-shot learning, have not been discussed in the recent \hs literature. \citet{goodfellow2016} defined zero-shot learning as an extreme form of transfer learning. Applying this concept to NLP, a model trained on one language or domain learns to predict samples from an unseen language using the latent structures of a pre-trained language model aligned across multiple languages. In cross-lingual few-shot learning, a percentage of samples from the target language is added to the training on the source language, thus, strengthening cross-lingual and task-specific alignment \citep{schuster2019cross}. 

Cross-lingual approaches are expected to bridge the deep learning performance gap between languages that have large corpora available and low-resource languages \citep{adams2017cross} and have been named as a hot topic for the next ten years by \citet{zhou_duan_wei_liu_zhang_2018}. One possible reason for the lack of application of these techniques to hate speech is a lack of appropriate, publicly available data sets. An additional problem is the varying definition of hate speech used in different data sets, preventing the combined use of any given high-resource language corpus with any given low-resource corpus. Here, we use the data set from the ACL 2019 Semantic Evaluation challenge (SemEval) Task 5 \citep{basile-etal-2019-semeval} that contains both English \textbf{(EN)} and Spanish \textbf{(ES)} \hs tweets, identified according to the same definition of \hs targeting women and immigrants, to develop monolingual and cross-lingual models.

As \hsd is already very challenging to model in an uni-language setting \citep{macavaney2019hate, zhang2019hate, fortuna2018survey}, most existing work in this and related fields \citep{Benballa2019,  aggarwal2019ltl, Zhou2019, Pelicon2019, Pavlopoulos2019, Wu2019a, Liu2019} focuses on enhancing NLP state-of-the-art deep learning architectures, namely variations of Transformer Language Models (TLM), such as bidirectional encoder representations from Transformers (\bert) \citep{bert} or cross-lingual language model pre-training (\xlm ) \citep{xlm}. The majority of SemEval submissions that successfully detected and classified offensive language on social media (Task 6) utilised \bert variations, fine-tuning them in an end-to-end fashion. 

In this paper, we describe a novel approach for \hsd that uses frozen TLM architectures to extract features -- the text representations -- only from some of the TLM layers. By using TLMs purely as a feature extractor, we avoid the computation expensive, task-specific, fine-tuning training step, which adjusts up to 8.3 billion trainable parameters \citep{shoeybi2019megatron}. This strategy is briefly mentioned in the original \bert paper \citep{bert}, but only \citet{Peters2019} have carried out a broader evaluation. To the best of our knowledge, our approach is a novel idea both in the context of cross-lingual learning as well as for \hsd with small data sets.

Following the extraction of the representations, we feed them into 12 different classification blocks of varying complexity, which we install as trainable layers on top of the feature extraction network. Inspired by the results and intuition of state-of-the-art computer vision attention blocks, we also step-wise derived and crafted a novel representation classification block, Attention-Maximum-Average Pooling (\axel), for this particular task. 

During our investigations, we noticed that good recall performances often resulted in poor precision and an unstable F1 for EN. We attribute this to an out-of-domain sampling and propose reshuffled partitions of the English data \textbf{(EN-S)}. All experiments were performed on both the original and proposed split. All associated code is openly available\footnote{Access on github.com/username/projectname}. 

Our contributions are as follows. Firstly, we demonstrate that frozen TLMs can serve as pure deep feature extractors for \hsd that only need a fraction of trainable parameters compared to the normal fine-tuning approach. Secondly, we propose a novel classification block \axel that enabled competitive results on uni-language and cross-lingual \hsd. Thirdly, we demonstrated the efficiency of zero- and few-shot learning in this setting and, finally, we identify serious limitations in the generalisability of models trained with the EN HatEval data and propose a new sampling.

\section{Network Architecture Components}
\begin{figure*}[ht]
    \centering
    \includegraphics[width=1.0\textwidth, trim={1.2cm 11.8cm 2.7cm 12cm},clip]{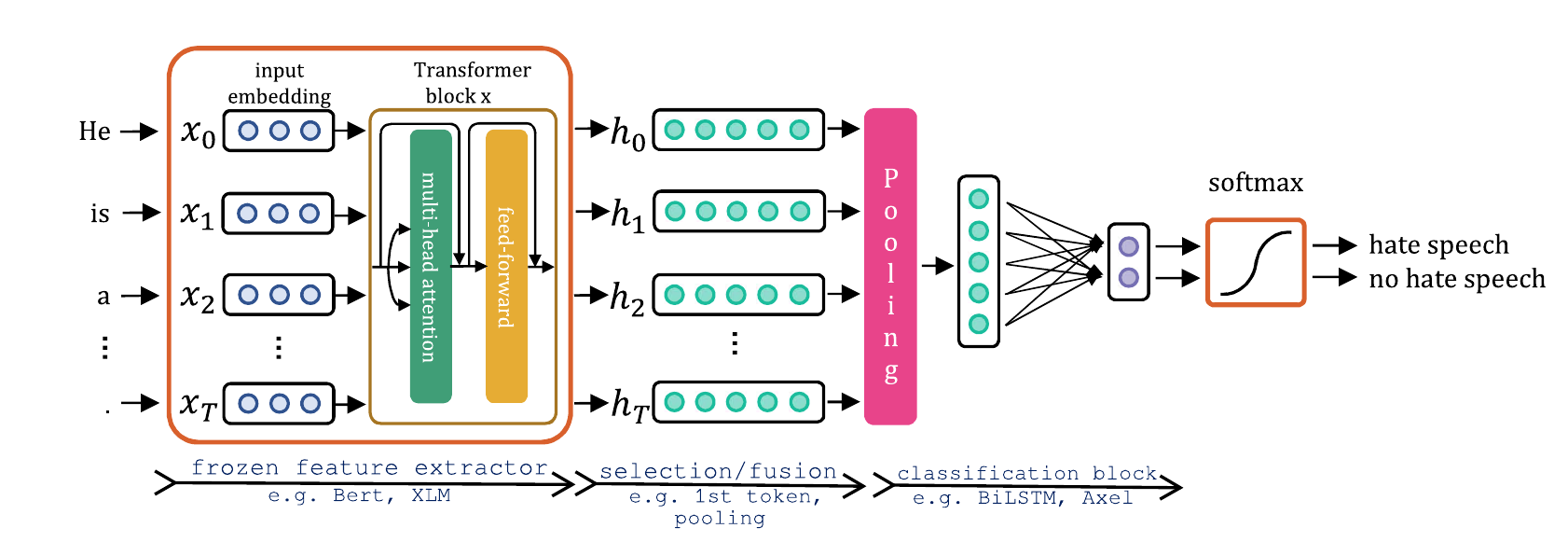}  
    \caption{Exemplified illustration of the architecture (\eg \texttt{BERT\_AvgPool}). The frozen \bert model acts as a feature extractor (brown frame). Next, a module selections one or many output representations \eg by pooling over the sequence (pink). The output of the selection is used as input for another representation enhancement module \eg a dense feed-forward layers (green).}
    \label{fig:overall}
\end{figure*}

The high-level structure of our architecture is depicted in Figure \ref{fig:overall}. Initially, the network receives the language tokens of the input text. This sequence is propagated forward through a frozen TLM architecture (either \bert or \xlm), extracting the deep language features. In the next step, some or all of the extracted representations are selected or fused for further processing. For example, this can be the last representation of the output layer for a pure one dimensional classification block or the entire sequence for a sequential compression. Next, these representations are fed into a classification block where the target is predicted.

We also replaced the frozen TLM by common word embeddings, such as, \fast \citep{grave2018learning}, \glo \citep{Pennington14glove} or \fasta \citep{joulin2018loss} to encode the input text. \fast showed better results than \glo on Twitter \hs data \citep{van2018challenges}. \fasta has multi-language capabilities. The encoded representation was fed into a single or double recurrent neural network layer, \eg a Long short-term Memory Network \citep{hochreiter1997long} or its bidirectional version (BiLSTM), which have previously proved more effective than Convolutional Neural Networks for \hsd \citep{rizos2019augment, van2018challenges}.
\subsection{Extracting Transformer Language Model Features}
We do not perform a fine-tuning step, instead we use the TLMs as a frozen language feature extractor. For a general description of the Transformer architecture we refer to \citet{vaswani2017attention}.

\bert was the first TLM to successfully train text representations bidirectionally \citep{bert}. Since a Spanish version of BERT\textsubscript{LARGE} was not available, we used the multilingual cased BERT\textsubscript{BASE} as one of our feature extractors. \bert is not explicitly cross-lingual pre-trained, whereas \citet{xlm} aligned the language representations in a two-step pre-training process for \xlm. 
In the first step, a masked language modeling is trained unsupervised using byte pair tokenised sub-words \citep{sennrich2015neural}. In the second step, translation language modeling uses sentence pairs from different languages and feeds them in parallel into the model. 
\subsection{State-of-the-art Computer Vision Classification Blocks and Derived \axel}\label{sec:sotacv}
Our novel classification block was designed to efficiently condense task-specific representations from a sequence of context-specific, general text representations from a general TLM. To do this, we analysed the structure of recent state-of-the-art attention modules, CAB \citep{rcab}, CBAM \citep{cbam}, CSAR \citep{csar}, and RAM \citep{ram} that can simultaneously compress and enhance feature representations, \eg in image super-resolution tasks. Most separate the attention block into spatial and channel attention layers. They extract information across the filter dimensions, then capture inter-dependencies in the feature channels utilising a three-step squeeze, excitation and scaling procedure \citep{ram}. 
Since in our case the input data have one dimension less, specifically the RGB channels, we adapted the modules for text representation compression. For the spatial attention, we utilised a 1D convolution operation over the entire sequence length to combine all representations. For the channel attention, we used pooling over the feature vector dimension to distill information from each individual sequence representation.   

\begin{figure}[!htp]
    \centering
    \includegraphics[width=0.5\textwidth, trim={1.3cm 11cm 1.1cm 6.5cm},clip]{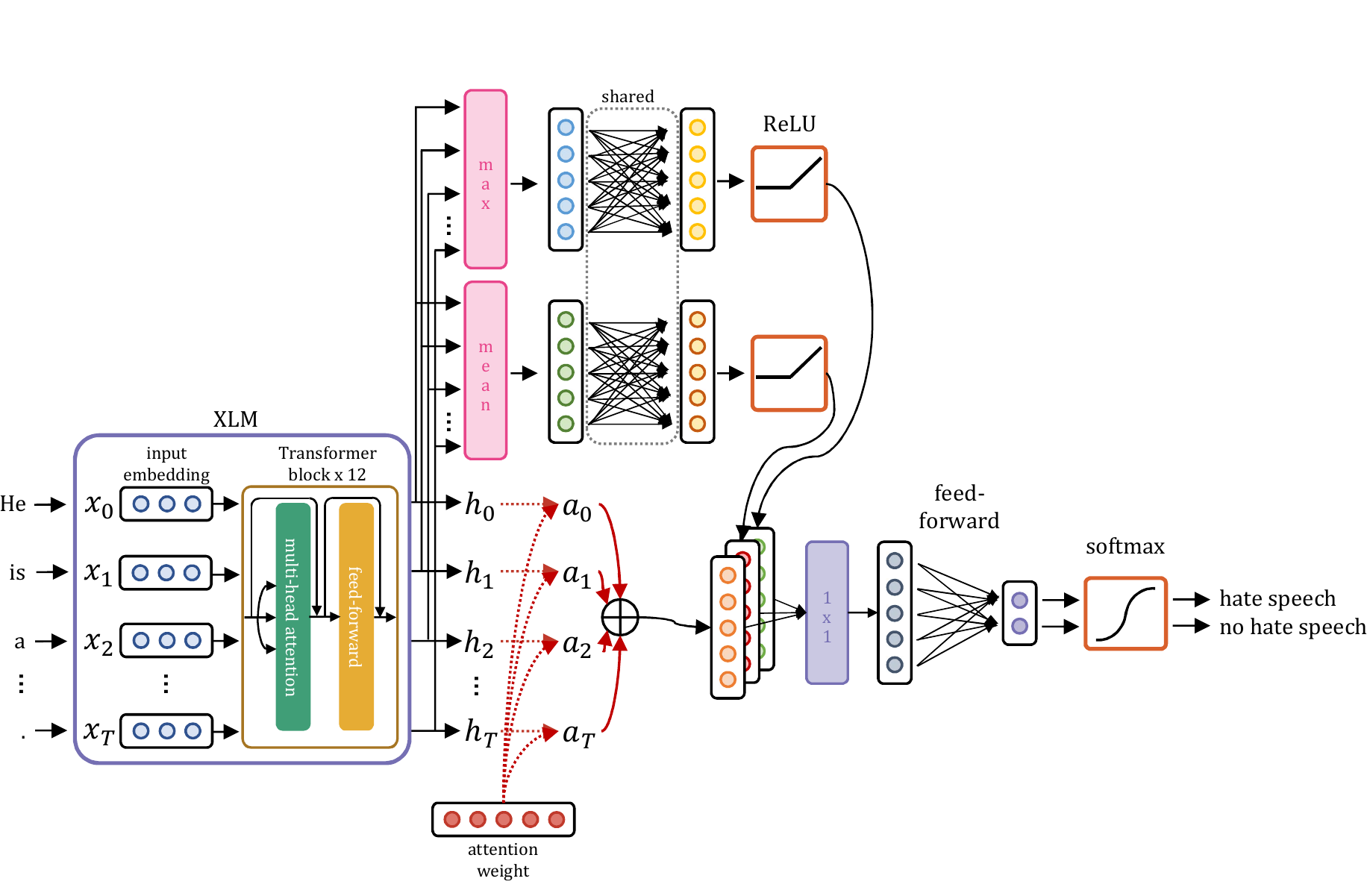}
    \caption{The \wip{} classification block compresses and enhances \xlm features. Similar to a channel attention modules, a maximum and average pooling on the \xlm output is used, followed by a feed-forward layer with shared weights and non-linear ReLU activation. This is fused with a context attention module with stacked feature maps as synthetic filter channels. The filters are fused by a one-dimensional convolution (denoted 1x1).}
    \label{xlm_wipname}
 \vspace{-10pt}
\end{figure}

We combine the most promising modules step-wise to create \axel (\cf Figure \ref{xlm_wipname}). The context and two different channel attention modules enhance the underlying \xlm features. Sharing the weights between the two different channel attention modules results in more robust representations, while the subsequent ReLU adds additional non-linearity. The two resulting representations are then fused with the output of a context attention module by stacking the three feature maps as synthetic filter channels. Next, a one-dimensional convolution (denoted 1x1) is used to deeply fuse the stacked filters. Finally, a feed-forward layer with softmax activation enabling \hs prediction.

\section{Data}
\subsection{HatEval Data Set}
\begin{figure}[b]
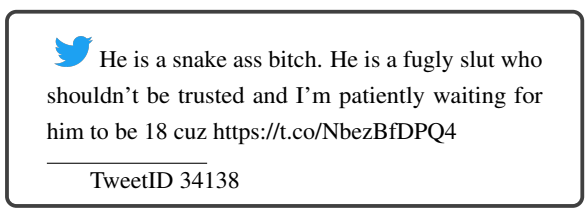

    \tweet{{\small He is a snake ass bitch. He is a fugly slut who shouldn't be trusted and I'm patiently waiting for him to be 18 cuz… https://t.co/NbezBfDPQ4}\blfootnote{TweetID 34138}} 
   \caption{Example tweet declared as non-hateful but clearly hateful against men.}
  \label{fig:hatefulnothateful} 
\end{figure}

We evaluated the effectiveness of our proposed \hsd models on the HatEval data set released as part of the SemEval task 5 \citep{basile-etal-2019-semeval,SemEvalProc2019} and focused on the first sub-task only. The data set definition aims at women and immigrants \hs. Hate speech directed at other groups, \eg men was labelled as not hateful (\cf Figure \ref{fig:hatefulnothateful}).

The dataset comprises around 13,000 EN tweets and 6,600 ES tweets. We performed a simple descriptive statistical analysis to verify there were no obvious set or label-related patterns in the data. The data are slightly imbalanced; 42\% were labelled as \hs and 58\% labelled as containing no \hs. A detailed analysis of the \hs text properties can be found in the appendix.


 
\subsection{Proposing New Partitioning of the English HateEval}
An analysis of the challenge baselines \citep{basile-etal-2019-semeval}, submissions, and our own initial tests demonstrated a large discrepancy in performance between EN and ES. The challenge submissions and baselines produced average F1 scores of 44.84\% and 68.21\% with EN and ES respectively, with no explanation provided in the subsequent retrospective account of the challenge \citep{basile-etal-2019-semeval}. Consequently, we investigated EN more closely and re-partitioned it before proceeding further. 
\subsubsection{Error Analysis of the English Sub-set: Out-of-Domain Sampling}\label{sec:error}

To investigate the low performances obtained using EN, we trained a simple baseline model using \fast embeddings and a BiLSTM. As expected, the model, considerably trained and tuned well on the validation partition, achieved on testing a low precision of 43\%, a high recall of 94\%, and resulted in 1,564 false positives in 2,971 test samples. 

\begin{table}[ht]
    \centering
    \setlength\tabcolsep{4.5pt} 
    \begin{tabular}{lcc}
        \hline 
        \textbf{key phrases} & \textbf{train+val} & \textbf{test} \\ 
        \hline
        build * wall & 97\%  & 35\%   \\ 
        MAGA & 88\%  & 29\%   \\ 
        illegal aliens & 89\%  &  33\%   \\ 
                               \hline
        total anti-immigration                         & 92\%  &  34\%   \\  \hline\hline   
        total anti-women & 79\%  & 46\%   \\ 
        \hline
    \end{tabular}
    \caption{Occurrence of discriminate phrases, grouped as anti-immigration and anti-women (\textquote{bitch}), associated to \hs on the training (train) and validation (val) partitions versus the test partition. The percentage is the \hs ratio, the number of \hs samples including a particular phrase divided by the total number of samples containing that phrase. \textquote{*} denotes that/the. MAGA = Make America great again. \label{tab:hate_speech_with_hashtag} }
\end{table}

\begin{figure*}[ht]
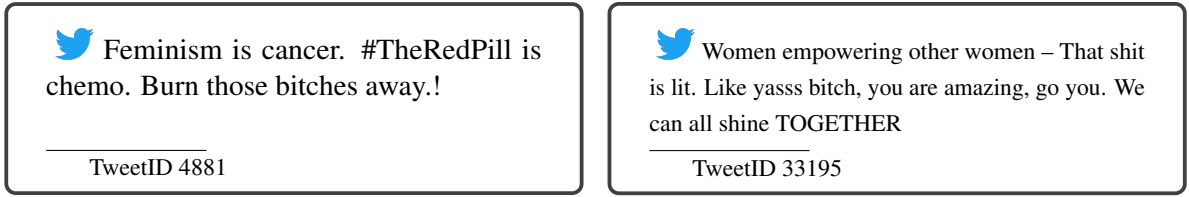
 
 \begin{minipage}{.49\textwidth}
  \tweet{Feminism is cancer. \#TheRedPill is chemo. Burn those bitches away.!\\ \blfootnote{TweetID 4881}} 
 \end{minipage}
 \begin{minipage}{.49\textwidth}
  \tweet{{\small Women empowering other women -- That shit is lit. Like yasss bitch, you are amazing, go you. We can all shine TOGETHER} \blfootnote{TweetID 33195}} 
 \end{minipage}
   \caption{Typical examples including the keyword \textquote{bitch} from training set (left) and test set (right). \label{fig:bitch_train}}
  
\end{figure*}

When we made spot checks of false positive samples, certain repeatedly occurring signal words stood out. The phrases \textquote{bitch}, \textquote{build the [or that] wall}, \textquote{make America great again} and \textquote{illegal aliens} occurred at least once in 80\% of false positives in the test set. Table \ref{tab:hate_speech_with_hashtag} compares context and corresponding labels. While training and validation sets were very homogeneous, the test partition has very different properties. We hypothesise that most deep learning algorithms would learn these phrases as discriminative linguistic markers based on the training and validation set, which then appear in the test set in a different context and cause the high number of false positives. In an example, Figure \ref{fig:bitch_train} shows a pair of tweets containing the word 'bitch'; one is from the training/validation set (left) and the other is from from the test partition (right).

In summary, we believe that out-of-domain sampling of the EN test set hinders the development of sensible models that behave similar on all partitions and drawing of meaningful conclusions from qualitative analyses (e.g. model error analysis). We speculate that the test set was not collected nor partitioned with the rest of the data and that different criteria were applied, be they thematic or temporal, so that the domain of data distribution differs widely. 
 
\subsubsection{Proposed New English Partitions}\label{sec:proposed}
Having identified a potential source of the deviations, we propose a simple approach to solve this issue. The goal of the new partitioning is an approximately equal distribution of data properties with the previously specified key phrases taking into account the binary label for \hs. This results in three categories: no key phrases, one or more anti-immigration key phrases and anti-women key phrases (\textquote{bitch}) -- for hateful and not hateful data points, leading to a total of six classes for partition stratification. We merged all EN partitions, then equally re-distributed the tweets according to six categories. In the end, the \hs ratio was balanced enabling a fair and comprehensible learning of automated \hsd models -- comparable to that of the Spanish subset. In general, such heavy effects of sampling indicates that the amount of data is not enough to learn discriminative, fully generalisable features for any ambiguous context.
\subsection{Preprocessing}
We cleaned the tweets from the HatEval dataset to avoid biased training influences \citep{Hassan2013}, providing a description of our procedure in the appendix. The process markedly improved unique word coverage using \fast word embeddings, from 44.06\% to 82.41\% for EN and from 55.93\% to 90.05\% for ES words. In regard to the full text coverage, we achieved 97.43\% for the EN and 98.08\% for ES tweets, demonstrating the effectiveness of our comprehensive procedure. 
\section{Experiments}
\subsection{Experiment settings}
The models were implemented in Python 3.6 using PyTorch. Without fine-tuning, we could use a moderate hardware (NVIDIA Tesla K80). All models use a cross-entropy loss function, Adam optimiser and early stopping. In addition, hyperparameters settings (\textbf{P}) of the abbreviation are provided in the appendix. We trained the models on training partitions measuring accuracy, precision, recall, and F1, but only report the latter for conciseness. 

\subsection{Baselines}
The challenge organisers provided two baselines \citep{basile-etal-2019-semeval}, a Most-Frequent-Classifier (\texttt{MFC}, EN: 36.7, ES: 45.1 in \% F1) and a Support Vector Machine (\texttt{SVM}) using tf-idf vectorisation (\texttt{SVM}, EN: 45.1, ES: 70.1 in \% F1). We have created additional baselines (Table \ref{tab:embedding_comparison_english}), as our data pre-processing differs to \citet{basile-etal-2019-semeval}. Also, we require a benchmark for our newly crafted EN-S partition. Finally, no deep learning baseline was provided in the challenge and comparing the new context to the conventional word embeddings seems relevant.

\begin{table}[ht] 
    \centering
    \caption{Performance comparison of our baselines, based on a \texttt{SVM} similar to the original (\texttt{Base\_SVM}), with a slightly tuned $C$-value ($3.5938$), and a BiLSTM with a feed-forward layer and 300 dimensional word embeddings, namely, \glo (\texttt{Base\_GV}), \fast (\texttt{Base\_FT}) and \fasta (\texttt{Base\_FTA}). We report the F1 in \%.} 
    \setlength\tabcolsep{5.5pt} 
    \begin{tabular}{lrrr}
         \toprule
          \textbf{Model} & \textbf{EN}  &  \textbf{EN-S} &  \textbf{ES} \\ 
        \midrule
        \texttt{Base\_SVM} & 59.78 & \textbf{65.43} & 64.90\\
        \texttt{Base\_GV} & 58.93  & 63.44 & 66.14\\
        \texttt{Base\_FT} &  \textbf{60.18}  & 61.75 & \textbf{67.49}  \\
        \texttt{Base\_FTA} & 58.08 & 58.19  & 63.63 \\ 
        \bottomrule
    \end{tabular}
    \label{tab:embedding_comparison_english}
\end{table}

On the EN-S data set, our \texttt{Base\_SVM} achieved strong F1 scores with well balanced sub-metrics. We speculate that the performance of the word embeddings may be directly related to the amount of training data used in the embedding training. This behaviour is analogous to the results on the Spanish data set, where \texttt{Base\_FT} achieved the best results. The F1 of the SVC baseline is above the average result of the challenge participants \citep{basile-etal-2019-semeval}, and can, therefore, be considered a strong entry-level baseline. Overall, we have trained strong and robust baselines for both languages and evaluated various word embeddings. Based on these results, we choose \texttt{Base\_FT} as our main deep learning baseline.
\subsection{Results}
\subsubsection{Viability of Transformers as Deep Feature Extractors for Uni-Language Hate Speech Detection}

\paragraph{\bert base} 
We started with naive approaches utilising classification blocks with no or few trainable parameters. \citet{bert} extracted the first token of the final \bert layer and fed them into a softmax layer (\texttt{Bert1LT\_Dense}). We also examined if the most informative token could be learnt by applying a global max-pooling layer ( \texttt{Bert1L\_MaxPool}) over the temporal sequence to limit the computational costs and memory footprint plus provides translation invariance. \texttt{Bert1L\_AvgPool} with average pooling enabled us to evaluate if the the average representation is informative. As evident in Table \ref{tab:bertdd}, all models perform worse than our baseline, with the pooling solutions performing better than the recommended first token approach.
 
We utilised the entire sequence of \bert by feeding it into a double stacked BiLSTM (\texttt{Bert1L\_2LSTM}) similar to \citet{bert}, where the sequence outputs of the last four layers are extracted, concatenated across the layers, and fed into a two-layer BiLSTM (\texttt{Bert4L\_2LSTM}). The step-wise encoding of all the tokens by an LSTM increases the prediction performance considerably, on the EN-S data set by more than 7\% (\texttt{Bert1L\_2LSTM}) and ES still marginally better than our benchmark. The previous experiments used the final \bert layers. Thus, an entire forward pass through the network was necessary. We evaluated the feasibility to extract only the first (\texttt{Bert1F}). For the EN-S data set, both achieved results above the benchmark and only slightly worse than propagated till the final layer. 

\begin{table}[ht]
    \centering
    \caption{Performance of multiple models: \texttt{Bert1L\_ Dense} uses only the first token and a dense layer, \texttt{Bert1L\_AvgPool} uses average pooling, and \texttt{Bert1L\_MaxPool} max pooling over the last \bert layer. \texttt{Bert1L\_2LSTM} uses one and \texttt{Bert4L\_2LSTM} of the last \bert layer and feed them into an BiLSTMs, while \texttt{Bert1F\_1LSTM}, or two \texttt{Bert1F\_2LSTM} extract the first \bert layer. Analogous the XLM models. The models are compared on the English (EN), English reshuffled (EN-S), and Spanish (ES) data set and all results are reported in F1 \%.
    }
    \setlength\tabcolsep{4pt} 
    \begin{tabular}{lcccc}
         \toprule
         \textbf{Model}  &  \textbf{P} & \textbf{EN}  &  \textbf{EN-S} &  \textbf{ES} \\ 
        \midrule
        \texttt{Base\_FT} & A & 60.18 &  61.75 & 67.49 \\
        \hline \hline
        \texttt{Bert1L\_Dense} & F &  56.81 & 61.31 & 51.43 \\
        \texttt{Bert1L\_MaxPool} & G & 59.86 & 62.64 & 56.62 \\
        \texttt{Bert1L\_AvgPool} & G & 58.81 & 58.54 & 59.74 \\
        \midrule
        \texttt{Bert1L\_2LSTM} & D & 60.55 & \textbf{69.04} & 65.23 \\
        \texttt{Bert4L\_2LSTM} & D& \textbf{61.00}  & 68.98 & \textbf{67.57} \\
        \midrule
        \texttt{Bert1F\_1LSTM}& F & 59.37 & 67.75 & 64.85 \\
        \texttt{Bert1F\_2LSTM} & F & 59.21  & 67.28  & 64.09 \\
        \midrule \hline
        \texttt{XLM1L\_Dense} & G &  60.89 &  \textbf{67.73} & 64.75 \\
        \texttt{XLM1L\_2LSTM} & H & 60.20  &  59.27 & 62.37 \\
        \texttt{XLM1L\_Att} & G & \textbf{61.61} & 67.56 & 62.33 \\
        \bottomrule
    \end{tabular}
    \label{tab:bertdd}
\end{table}

\paragraph{\xlm base}
\xlm is designed to train cross-lingual TLMs and is the foundation for our zero- and few-shot approaches. We transferred the block design one to one from \bert. Noticeable is the strong performance of \texttt{XLM1L\_Dense} on all three sub-sets and the weak performance of \texttt{XLM1L\_2LSTM} (\cf Table \ref{tab:bertdd}). This also stands in contrast to the \bert model, where the sequential models were clearly superior to the non-sequential models. Given that sequential encoding does not add value to the use of \xlm tokens, we also tried to learn a weighted representation through an attention layer \texttt{XLM1L\_Att}. However, the model showed great performance on the EN set, but cannot beat \texttt{XLM\_Dense} on the EN-S and ES.
 
\subsubsection{Analysis of Advanced Classification Block Designs: \axel}

For the development of the novel \axel classification block, we got inspired by the latest attention developments in computer vision. Table \ref{tab:xlm_attention_adaption_performance} illustrates the performance of these blocks, whereby \texttt{XLM\_RCAB} performs best on the EN and ES. Looking at the architectures in detail, it is the only one which does not utilise spatial attention. It can be deduced that, spatial attention is not ideal for our purpose, while the for text adjusted channel attention adds value to the representations. Furthermore, it is evident that the newly proposed \axel classification block produced by far the best result, exceeding all other adapted blocks by at least 7\% F1 on the EN-S and more than 2\% for the ES. We provide an ablation study of all \axel components in the appendix.

\begin{table}[ht]
    \centering
    \caption{
    Comparison of \xlm classification blocks: RCAB \citep{rcab}, CBAM \citep{cbam}, CSAR \citep{csar}, RAM \citep{ram}, and our newly developed \axel. All results are reported in F1 \%.
    }
    \begin{tabular}{lcccc}
         \toprule
         \textbf{Model}  &  \textbf{P} & \textbf{EN}  &  \textbf{EN-S} &  \textbf{ES} \tabularnewline 
        \midrule
         \texttt{XLM\_RCAB} & K & \textbf{62.36} & 61.65  & 60.28 \tabularnewline
         \texttt{XLM\_CBAM} & F & 60.90 & 59.67 & 54.25 \tabularnewline
         \texttt{XLM\_CSAR} & M & 61.45 & 63.85 & 50.17 \tabularnewline
         \texttt{XLM\_RAM}  & M & 60.30 & 60.67 & 55.21 \tabularnewline
        \midrule
        \wip{} & F & 62.03 & \textbf{71.16} & \textbf{69.70} \tabularnewline
        \bottomrule{}
    \end{tabular}
    \label{tab:xlm_attention_adaption_performance}
\end{table}

\subsubsection{Cross-lingual learning} \label{sec:cross}

\paragraph{Zero-shot learning}
To evaluate zero-shot capabilities, the models are tuned on the training set of one language and evaluated in another. Table \ref{tab:zeroshot_xlm} illustrates that \wip achieved the best results except on the EN data set, where \texttt{XLM\_Dense} performed best but suffered one more time from high false positives. Overall, the models show general learnability, however, lack generalisability and the performance is much worse than in our monolingual experiments. 

To probe if the causes of the performance loss are either on the data or the model side, we carried out additional experiments. To rule out that the different nature of the EN and ES training and test partitions is the reason, we automatically translated\footnote{https://aws.amazon.com/translate/} the test set into the language of the training set. The improved results in Table \ref{tab:zeroshot_xlm} indicate that the partition composition is probably not the cause, leaving only the extracted latent representations, which seems either not perfectly aligned for our task \hs or across languages.

\begin{table}[ht]
    \centering
    \caption{Zero-shot performance comparison of \xlm base \texttt{XLM\_Dense(G)}, \texttt{XLM\_Att(G)}, and \wip{} \texttt{(F)} across languages and predictions on the translated test set in F1 in \%. This shows that the latter outperforms other \xlm based models as well as that the output latent structure for various languages differs. $train \Rightarrow test$; $original \rightarrow{} translation$. }
    \begin{tabular}{llrrr|r}
         \toprule
         & \texttt{Dense} & \texttt{Att} & \texttt{AXEL} 
        \\ \midrule
        \shortstack[l]{EN$\Rightarrow$ES} & 41.31 & 34.37 & \textbf{53.42} \\
        \shortstack[l]{ES$\Rightarrow$EN} & \textbf{60.83} & 48.47 & 52.48 \\

        \shortstack[l]{ES$\Rightarrow$EN-S} & 49.38 & 39.10 & \textbf{53.24} \\
        \midrule
        \shortstack[l]{EN$\Rightarrow$(ES$\rightarrow$EN)} & 60.59 & 62.40 & \textbf{64.39} \\
        \shortstack[l]{ES$\Rightarrow$(EN$\rightarrow$ES)} &  56.89 & 49.17 & \textbf{58.31} \\
        \shortstack[l]{ES$\Rightarrow$(EN-S$\rightarrow$ES)} & 56.57 & 49.17 &  \textbf{65.04} \\
        \bottomrule
    \end{tabular}
    \label{tab:zeroshot_xlm}
\end{table}

Furthermore, we tried to learn models in similar fashion using \fasta embeddings combined with various blocks. Most classifiers generalised badly, resulting in unstable losses (see appendix for experiments).

\paragraph{Few-shot learning}
The previous experiments have shown that zero-shot learning works, but performed much poorer than the monolingual models. Therefore, we determined whether the extracted latent structure can be stronger aligned or learnt by the classification block, when we inject a few percent of the samples of the predictive language into the training set.

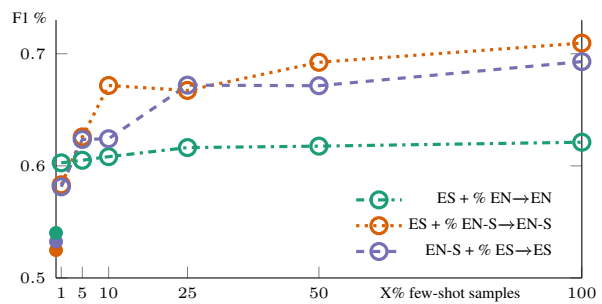
\begin{figure}[!ht]
    \centering
        \begin{tikzpicture}[baseline]
        \begin{axis}[
            axis x line* = bottom, axis y line = box,
            y label style={at={(axis description cs: -0.05,0.95)},font=\tiny, rotate=270,  anchor=south},
            x label style={at={(axis description cs: 0.75,0.005)}},
            ylabel={F1 \%},
            xlabel={X\% few-shot samples},            
            label style={font=\tiny},
            tick label style={font=\tiny},
            xmin=0, xmax=100,
            ymin=0.5, ymax=0.73,
            xtick={1,5,10,25,50,100},
            ytick={0.5,0.6,0.7,0.8},
            cycle list/Dark2,
            width=8.5cm,height=5cm,
            legend style = {draw=none,fill=none, font=\tiny, scale=0.5},
    	    legend cell align={right},
    	    legend pos= south east,
            ymajorgrids=false,
            grid style=dashed,
            every axis plot/.append style={very thick}
        ]
         

        \addplot+[dashdotted,mark=o,mark options={scale=1.5,solid}]
            coordinates {
            (1,0.6027) (5,0.6050) (10,0.6082) 
            (25,0.6163) (50,0.6176) 
            (100,0.6211)
            };
            \legend{ES + \% EN$\xrightarrow{}$EN}
        \addplot+[mark=*, very thick, Dark2-B, forget plot]
            coordinates {
            (0,0.5248)
            };

        \addplot+[dotted,mark=o,mark options={scale=1.5,solid}] 
            coordinates {
            (1,0.5831) (5,0.6258) (10,0.6717) 
            (25,0.6673) (50,0.6924) 
            (100,0.7095)
            };
            \addlegendentry{ES + \% EN-S$\xrightarrow{}$EN-S}
        \addplot+[mark=*, mark options=solid, very thick, Dark2-C, forget plot]
            coordinates {
            (0,0.5324)
            };
        
        \addplot+[dashed,mark=o,mark options={scale=1.5,solid}]
            coordinates {
            (1,0.5813) (5,0.6236) (10, 0.6242) 
            (25, 0.6720) (50, 0.6715) 
            (100, 0.6930)};
            \addlegendentry{EN-S + \% ES$\xrightarrow{}$ES}
        \addplot+[mark=*, very thick, Dark2-A, forget plot]
            coordinates {
            (0,0.5402)
            }; 
         
        \end{axis}
        \end{tikzpicture}
     \caption{Few-shot performance in F1 of the \wip{} over percentage of added few-shot samples (detailed table in appendix). The zero-shot results are shown with a filled circle. We added a certain percentage of the evaluation-language training set to the source-language training set. 0\% added equals the zero-shot training case. The results show that adding 1\% substantially improved to zero-shot. \label{fig:few-shot}}
\end{figure}

Figure \ref{fig:few-shot} shows that injecting only 1\% of the data led to a boost in performance of almost 5\% F1. Subsequently, the training continues mostly as expected, improving incrementally with more data. The final result for the EN-S exceeded even the monolingual experiments, which we attribute to data augmentation by the additional injected data. In order to verify this idea, we additionally experimented with augmentation techniques, such as, translation chains, but were unsuccessful in demonstrating an overarching and clearly positive effect. The original EN shows an atypical behaviour, and apparently barely improving over time due to the out-of-domain sampling issue (\cf section \ref{sec:error}).

In order to verify that the network has actually learnt cross-lingually and not only from the few injected samples, the same experiments were carried out using exclusively the few-shot samples, thus, excluding the full training-language. These clearly show that the model makes use of cross-lingual structures. For example, at 10\% few-shot training samples, we obtained results of only 58.10\% (EN), 59.06\% (EN-S), and 58.66\% (ES) F1, which are clearly lagging behind the few-shot results for EN-S and ES.

\subsection{Discussion}
One central motivation of our work was to access and improve cross-lingual learning, especially for low-resource languages. While the models using cross-lingual zero-shot learning produced mixed results, the benefits of few-shot learning based on extracted features are evident.  Unfortunately, we were unable to use a genuinely low-resource language because of the limited availability of multi-lingual \hs corpora using the same definition with one being of a rare language. However, using our artificially reduced high-resource data sets, a parallel training, even with very few data sets, resulted in a stronger alignment of \xlm representations.

Reducing computation during task-specific fine-tuning was another important motivator in this work. When we compare the architectures presented here with the fine-tuning approach, the \texttt{Bert4L\_2LSTM} classification block has less than 2M trainable parameters, while \texttt{XLM\_AXEL} has only around 1M. At the same time, we have to train 177M parameters for \bert, and 249M parameters for \xlm to train these networks end-to-end. This insight is valuable for academia or industry where less resources are available. This is of less significance in terms of the inference time due to the highly efficient architecture of Transformers \citep{peters2018dissecting}. 

However, while far less parameters are used, we demonstrated that our novel \axel classification block on frozen TLM could still easily beat our strong baselines. When ranked alongside the HatEval 2019 results, our approach ranked second on EN and is close to the third quartile (71.65\% F1) on ES.


\section{Related Work}

In the original paper, \citet{bert} included little information about using extracted \bert features for named-entity recognition. More extensive research was carried out by \citet{Peters2019}, evaluating both fine-tuned features as well as features from the general language model on a wider range of tasks. Predicting sentiment on movie reviews is the closest task to \hsd, showing that there is a slight trade-off between performance and computation cost for fine-tuning. Besides these works, there appears to be no other research conducted on small data sets or \hs data utilising TLMs as pure feature extractors, making it an interesting area to investigate.

As interest in the research community in \hsd has grown; the number of publicly available data sets has also increased. \citet{Waseem2016a} created and extended \citep{Waseem2016} an English \hs data set based on tweets. \citet{davidson2017automated} also focused on offensive language on twitter, while \citet{DeGibert2019} crawled the white supremacist website Stormfront. There are multiple data sets available in languages other than English including Italian \citep{Bosco2018}, Portuguese \citep{fortuna2019hierarchically}, and Indonesian \citep{ibrohim2019multi}. However, only \citep{basile-etal-2019-semeval} provides a corpus in two languages (English and Spanish) using the same definition of \hs, enabling us to tackle this topic from a multilingual perspective.

Recent detailed comparisons of traditional and deep learning approaches \citep{kshirsagar2018predictive,Robinson2018,van2018challenges,lee2018comparative,zhang2018detecting} have demonstrated superior performances by the later in \hsd. Transformers became especially popular for deeply modelling language, resulting in the quick and wide adoption of networks such as \bert \citep{bert} and \xlm \citep{xlm}. In Task 5 of SemEval-2019 for \hsd, the second-placed submission in the Spanish challenge \citep{Gertner2019} used a fine-tuned \bert by adding additional tweets to the corpus. Also in the closely related Task 6, in which offensive language was detected \citep{zampieri2019semeval}, six out of the top ten top submissions used \bert. It is noteworthy that all the published participants in these two challenge tasks who used a TLM architecture fine-tuned their architectures \citep{zhang2018hate, Gertner2019,Benballa2019,Siddiqua2019,aggarwal2019ltl,Zhou2019,Zhu2019,Nikolov2019,Pelicon2019}.

An extensive survey of cross-lingually word embeddings can be found in \citet{ruder2017survey}. \citet{grave2018learning} and \citet{muse} cross-lingual pretrained embeddings are widely used, likely because they are freely available. \citet{artetxe2018massively} suggested zero-shot, cross-lingual sentence embeddings, which showed a strong performance on some language combinations and partially competed with \bert. \citet{wu2019beto} attempted to adjust the Transformer for \bert for cross-lingual tasks, but \xlm performed superiorly. 

\section{Conclusion}
The detection of hate speech on social media platforms is vital to prevent the incitement of violence. A particular challenge is the development of reliable, automatic detection systems, where there is a lack of task-specific, low-resource language data.

The aim of this work was to evaluate TLMs as deep feature extractors for this task. First, we built strong baselines and assessed various classification blocks for the detection of uni-language \hs. The performance of the TLM-based models greatly surpasses that of those based on conventional word embeddings and demonstrated promising results compared to the submissions of challenge participants. On the EN, for example, they ranked second. Second, the poor generalization behaviour we observed on the EN partitions could be attributed to out-of-domain sampling and motivated our proposal of a newly stratified data split. We accompanied this with a first benchmark of EN-S partitions that we hope others will build upon, enabling sensible comparisons in the future. Finally, from our investigation of potential block designs, our results indicate two different strategies are required to successfully use \bert and \xlm representations. While \bert efficiently utilised the entire sequence of representations, \xlm worked better using only the first token. This motivated \axel, which is, derived from state-of-the-art computer vision modules, designed to extract a wide range of stable features out of one compressed representation. We artificially simulated low language resources to demonstrate the cross-lingual capabilities of our \axel module that outperformed our baselines by far and gave valuable insights for future research in this field.

Investigations of the representational differences from an architectural and training perspective (\textit{Why are \xlm representations less effective for sequential blocks?}) as well as general \axel capabilities are interesting future research directions.



\bibliography{acl2020}
\bibliographystyle{acl_natbib}

\newpage

\noindent \textbf{Appendices to accompany "Cross-lingual Zero- and Few-shot Hate Speech Detection utilising frozen Transformer Language Models and \axel"}

\appendix

\section{Data set}

\textbf{A.1 Data Analysis of Hate Speech Language Properties}.
\newline
In the Spanish corpus, we identified two overly long tweets (450 and 197 words) that contained obviously multiple, concatenated tweets. We removed these outliers. Also the tweets length seem to be quite homogeneous between the sets: labels and languages with around 22 words ($\pm$ 11) and approximately 140 chars ($\pm$ 70) in EN as well as roughly 21 words ($\pm$ 14) and almost 129 chars ($\pm$ 86) in ES. We also compared the usage of all caps words indicating emphasization or screaming and special characters (!, ?, \#, ., @), both possible identifiers for \hs. In the EN tweets, there is on average an almost 30\% increase in the usage of block capital written words in a \hs tweet (1.07 $\pm$ 2.56 vs 1.39 $\pm$ 3.29) while in Spanish this difference is insignificant (1.30 $\pm$ 4.19 vs 1.39 $\pm$ 4.68). In terms of the special characters, hateful EN tweets use exclamation marks nearly double as often than non-hateful tweets, also the usage of hashtags is slightly increased (0.92 vs 1.29 per tweet). Interestingly, this is contrary to the Spanish tweets, where the usage of hashtags is halved with 0.24 hashtags for non-hateful and 0.13 for hateful tweets, indicating cultural differences in the usage of hashtags. For all others properties we could not find clear differences. 

\smallskip

\noindent\textbf{A.2 Data Cleaning Procedure}.
\newline
We eliminated all mentions (\textquote{@}) since we expect that the mentioned usernames could be associated more closely to one of the classes. Similar hyperlinks might be biased or, in the case of shortened ones have no predictive value at all, and, thus, are also excluded. Apostrophes, for example, \textquote{Trump's wife} indicating genitives \textquote{'s} are completely removed, in singular form, this also included the latter.

The informal style of tweets makes it necessary to replace contractions by the full words (\textquote{you're}$\rightarrow$ \textquote{you are}). Besides a few simple transformations, for instance, standardising special characters (\eg \textquote{--} (en-dash) or \textquote{---} (em-dash) $\rightarrow$ \textquote{-} (hyphen), and numbers ("2nd" $\rightarrow$ second) also more complex ones were necessary. 
To reduce negative effects of colloquial word useage, we, first, transformed specific, rare but decisive words and abbreviations, such as, \textquote{MAGA} $\rightarrow$ \textquote{make America great again} or \textquote{Obamacare} $\rightarrow$ \textquote{Obama healthcare system}. Second, concatenated words and hash tags with capitalised letters (camel case) were separated to its unique words. Finally, we used the Python library emoji\footnote{\url{https://github.com/carpedm20/emoji/} (accessed 6 October 2019)} to replace smileys by text.

\section{Hyperparameter Abbreviation Table}
Table \ref{tab:hyperparameters} shows the non-static hyperparameters and values for each hyperparameter abbreviation (P).

\begin{table}[H]
    \caption{List of hyperparameter (P) combinations used.}
    \label{tab:hyperparameters}
    \centering
    \begin{tabular}{l|rrrr}
        \toprule
        \textbf{P} &  
        \multirow{2}{*}{\shortstack[r]{learning\\ rate}} &
        \multirow{2}{*}{\shortstack[r]{batch\\ size}} &
        \multirow{2}{*}{\shortstack[r]{RNN \\ feature size}} & \multirow{2}{*}{\shortstack[r]{RNN \\ dropout}}  \\ 
         &  \\ \midrule
         A & 0.001 & 32 & 128 & 0.0 \\
         B & 0.001 & 32 & 128 & 0.2 \\
         C & 0.0005 & 16 & 128 & 0.2 \\
         D & 0.00005 & 64 & 128 & 0.0 \\
         E & 0.00005 & 64 & -- & -- \\
         F & 0.0005 & 64 & -- & -- \\
         G & 0.00001 & 64 & -- & -- \\
         H & 0.0005 & 32 & 64 & 0.2 \\
         I & 0.0005 & 32 & 128 & 0.2 \\
         J & 0.0005 & 64 & 128 &  0.2 \\
         K & 0.00005 & 32 & -- &   -- \\
         L & 0.00005 & 32 & 128 &   0.0 \\
         \bottomrule
    \end{tabular}
\end{table}

\section{Additional Results Including Accuracy, Precision, Recall, and F1}

\noindent\textbf{C.1 Baselines}.

\begin{table*}[ht!]
    \centering
    \caption{Performance comparison of our baselines with all metrics reported in accuracy (ACC), precision (PRC), recall (REC), and F1 in \%. \texttt{Base\_SVM} is based on the original baselines, with  a  slightly  tuned C-value  (3.5938). The other models utilise 300 dimensional word embeddings, namely, GloVe (GV), fastText (FT) and, fastText aligned (FTA) and a BiLSTM with a single feed-forward layer. The models are compared on the English (EN), English reshuffled (EN-S), and Spanish (ES) data set.}
    \begin{tabular}{llrrr|r}
         \toprule
        Dataset & Model & ACC & PRC & REC & F1 \\ \midrule
        \multirow{4}{*}{EN}
        & \texttt{Base\_SVM} &  49.95 & 45.19   & 88.26 & 59.78 \\
        \cline{2-6} 
        & \texttt{Base\_GV} & 46.25 & 43.61 & 93.75 & 58.93 \\
        & \texttt{Base\_FT} &  \textbf{48.43} & \textbf{44.77} & \textbf{94.80} & \textbf{60.18} \\
        & \texttt{Base\_FTA} & 46.38 & 43.48 & 90.42 & 58.08 \\ \midrule
        
        \multirow{5}{*}{\shortstack[l]{EN-S}} 
        & \texttt{Base\_SVM} & 70.27 & 64.11 & 66.81 & 65.43 \\
        \cline{2-6} 
        & \texttt{Base\_GV} & 60.66 & 52.23 & \textbf{82.99} & \textbf{63.44} \\
        & \texttt{Base\_FT} &  64.72 & 56.17 & 70.37 & 61.75 \\
        & \texttt{Base\_FTA} & \textbf{65.42} & \textbf{59.11} & 59.09 & 58.19 \\ \midrule
        
        \multirow{5}{*}{ES}
        & \texttt{Base\_SVM} & 67.69 & 58.79 & 72.42 & 64.90 \\
        \cline{2-6} 
        & \texttt{Base\_GV} & 67.00 & 57.15 & \textbf{89.10} & 66.14 \\
        & \texttt{Base\_FT} &  70.31 & 61.02 & 77.04 & \textbf{67.49} \\
        & \texttt{Base\_FTA} & \textbf{70.69} & \textbf{64.91} & 64.01 & 63.63 \\
        
        \bottomrule
    \end{tabular}
    \label{tab:embedding_comparison_english_}
\end{table*}

The performance of the baselines including all metrics are given in Table \ref{tab:embedding_comparison_english_}. It demonstrates the imbalanced metrics of accuracy, precision, and F1 to the recall results.

\noindent\textbf{C.2 fastText aligned zero-shot learning}.

\begin{table*}[ht!]
    \centering
    \caption{Evaluation of the zero-shot performance of fastText aligned models. We trained a single LSTM layer \texttt{FTA\_LSTM(I)}, double LSTM layer \texttt{FTA\_2LSTM(I)}, and attention layer \texttt{FTA\_Att(I)} version. Many version were not efficiently trainable, even with early stopping, resulting in failing models indicated in \textit{italic}. The models are compared on the English (EN), English reshuffled (EN-S), and Spanish (ES) data set. Full results are reported in accuracy (ACC), precision (PRC), recall (REC), and F1 in \%.}
    \begin{tabular}{llrrr|r}
         \toprule
        Dataset & Model & ACC & PRC & REC & F1 \\ \midrule
        \multirow{3}{*}{\shortstack[l]{Train EN\\Test ES}}
        & \texttt{FTA\_LSTM}  & 39.62  & 	38.68  &  79.34  & 51.50 \\
        & \texttt{FTA\_2LSTM} & 39.37  & 	\textbf{39.20}  &  \textbf{85.60}  & \textbf{53.29} \\
        & \texttt{FTA\_Att} & \textbf{39.69}  & 	39.08  &  82.69  & 52.57 \\
        
        \midrule   
        
       \multirow{3}{*}{\shortstack[l]{Train ES\\Test EN}}
        & \textit{\texttt{FTA\_LSTM}}  & \textit{42.14}	 &  \textit{42.14}  &  \textit{100.00}  & \textit{58.73} \\
        & \texttt{FTA\_2LSTM} & 42.07  & 	42.10  &  99.80	  & 58.65 \\
        & \textit{\texttt{FTA\_Att}} & \textit{42.14}	 &  \textit{42.14}  &  \textit{100.00}  & \textit{58.73} \\
        
        \midrule   
        
       \multirow{3}{*}{\shortstack[l]{Train ES\\Test EN-S}}
        & \textit{\texttt{FTA\_LSTM}}  & \textit{42.11}	 &  \textit{42.11}  &  \textit{100.00}  & \textit{58.79} \\
        & \texttt{FTA\_2LSTM} & 42.09  & 	42.10  &  99.96	  & 58.77 \\
        & \textit{\texttt{FTA\_Att}} & \textit{42.11}	 &  \textit{42.11}  &  \textit{100.00}  & \textit{58.79} \\
        
        \bottomrule
    \end{tabular}
    \label{tab:zeroshot_fasttext}
\end{table*}

Aligned fastText allows cross-lingual applications. As depict in Table \ref{tab:zeroshot_fasttext}, most of the classifier based on aligned fastText word embeddings failed completely in \hsd .

\noindent\textbf{C.3 Few-short learning}.

\begin{table*}[!ht]
    \centering
    \caption{Results of \wip{} network over percentage of added few-shot samples. A certain percentage of the evaluation-language training set is added to the source-language training set. 0\% added equals the zero-shot training. The models are compared on the English (EN), English reshuffled (EN-S), and Spanish (ES) data set. Full results are reported in accuracy (ACC), precision (PRC), recall (REC), and F1 in \%.}
    \begin{tabular}{lrrrr|r}
         \toprule
        \multirow{2}{*}{Dataset} & \multirow{2}{*}{\shortstack[r]{\% few-shot \\ samples added}} & \multirow{2}{*}{ACC} &
        \multirow{2}{*}{PRC} & 
        \multirow{2}{*}{REC} & 
        \multirow{2}{*}{F1} \\ 
        &&&& \\ 
        
        \midrule   
        
        \multirow{5}{*}{\shortstack[l]{Train ES\\Test EN}}
        & 0  & \textbf{58.80} & \textbf{51.06} & 53.99 & 52.48 \\
        & 1  & 58.20 & 50.27 & 75.24 & 60.27 \\
        & 5  & 58.33 & 50.37 & 75.72 & 60.50 \\
        & 10 & 54.90 & 47.97 & 83.07 & 60.82 \\
        & 25 & 52.61 & 46.77 & \textbf{90.34} & \textbf{61.63} \\
        
        \midrule
        
        \multirow{5}{*}{\shortstack[l]{Train ES\\Test EN-S}}
        & 0  & 61.95 & 55.17 & 51.43 & 53.24 \\
        & 1  & 65.65 & 59.63 & 57.05 & 58.31 \\ 
        & 5  & 68.94 & 63.51 & 61.68 & 62.58 \\
        & 10 & 67.55 & 58.51 & \textbf{78.83} & \textbf{67.17} \\
        & 25 & \textbf{70.94} & \textbf{64.43} & 69.19 & 66.73 \\
        
        \midrule
        
        \multirow{5}{*}{\shortstack[l]{Train EN\\Test ES}}
        & 0  & 49.00 & 42.86 & 70.91 & 53.42 \\
        & 1  & 57.50 & 48.96 & 71.52 & 58.13 \\
        & 5  & 55.56 & 47.93 & \textbf{89.24} & 62.36 \\
        & 10 & 67.56 & 59.78 & 65.30 & 62.42 \\
        & 25 & \textbf{69.31} & \textbf{60.10} & 76.21 & \textbf{67.20} \\
        
        \bottomrule
    \end{tabular}
    \label{tab:few-shot_wipname}
\end{table*}

\begin{table*}[!ht]
    \centering
    \caption{Experiment to validate the cross-lingual learning by training \texttt{XLM1L\_AXEL(F)} purely on few-shot samples. The results indicate that \hs prediction cannot only be learnt with the few-shot samples, some models failed to train properly (\textit{italic}). The models are compared on the English (EN), English reshuffled (EN-S), and Spanish (ES) data set. Full results are reported in accuracy (ACC), precision (PRC), recall (REC), and F1 in \%.}
    \begin{tabular}{lrrrr|r}
         \toprule
        \multirow{2}{*}{Dataset} & \multirow{2}{*}{\shortstack[r]{\% few-shot \\ training samples}} & \multirow{2}{*}{ACC} &
        \multirow{2}{*}{PRC} & 
        \multirow{2}{*}{REC} & 
        \multirow{2}{*}{F1} \\ 
        &&&& \\ 
        
        \midrule   
        
        \multirow{3}{*}{EN}
        & 1 & 42.48 & 42.20 & \textbf{98.80} & 59.14 \\
        & 5 & 46.95 & \textbf{43.99} & 94.65 & \textbf{60.06} \\
        & 10 & \textbf{47.56} & 43.80 & 86.26 & 58.10 \\
        
        \midrule   
        
        \multirow{3}{*}{\shortstack[l]{EN-S}}
        & 1 & 45.27 & 43.22 & 95.49 & 59.51 \\
        & 5 & 43.09 & 42.52 & \textbf{99.82} & \textbf{59.63} \\
        & 10 & \textbf{68.37} & \textbf{64.91} & 54.18 & 59.06 \\
        
        \midrule
        
        \multirow{3}{*}{ES}
        & \textit{1} & \textit{41.25} & \textit{41.25} & \textit{100.00} & \textit{58.41} \\
        & \textit{5} & \textit{41.25} & \textit{41.25} & \textit{100.00} & \textit{58.41} \\
        & 10 & \textbf{48.56} & \textbf{43.88} & 88.48 & \textbf{58.66} \\

        \bottomrule
    \end{tabular}
    \label{tab:few-shot-crosslingual-learning}
\end{table*}

Extensive results of our few-shot approach as described in section \ref{sec:cross} are provided. We evaluated \wip{} for different percentages of injected few-shot samples in the training set (\cf Table \ref{tab:few-shot_wipname}) and ensured that the few-shot improvement is coming from learning on the original training samples -- not purely from the few-shot samples (\cf Table \ref{tab:few-shot-crosslingual-learning}.

\noindent\textbf{C.4 Ablation study for \texttt{XLM1L\_AXEL}}.
Table \ref{tab:xlm_AXEL_ablation} lists full results in accuracy (ACC), precision (PRC), recall (REC), and F1 in \% corresponding to the \axel elements. 

\begin{table*}[!ht]
    \centering
    \caption{Comparing the performance of \axel module by removing parts of it to determine the contributing factors: leaving out the max-pool module (\texttt{XLM1L\_AttAvgFC}), leaving out the avg-pool module (\texttt{XLM1L\_AttMaxFC}), not sharing weights (\texttt{XLM1L\_AttAvgFCMaxFC}),  aggregating the submodules instead of convolving (\texttt{XLM1L\_AttAvgFCMaxFCSum}), using a tanh activation function instead of ReLU (\texttt{XLM1L\_AttAvgFCMaxFCTanh}), adding an additional variance pooling submodule (\texttt{XLM1L\_AttAvgFCMaxFCVarFC}), the pure attention \texttt{XLM1L\_Att}, and \wip{}. Performed on the English, reshuffled English, and Spanish data set. Adding more submodules to \texttt{XLM1L\_Att} improves the performance, whereas the average pooling seems to have the strongest positive influence on the result.
    }
    \begin{tabular}{llrrr|r}
         \toprule
         & Model &  ACC & PRC & REC & F1 \\ \midrule
        \multirow{8}{*}{EN}
        & \wip{}                            & 51.53 & 46.30 & 93.93 & \textbf{62.03} \\
        & \texttt{XLM1L\_AttAvgFCMaxFCVarFC}  & 54.16 & 47.60 & 87.06 & 61.55 \\
        & \texttt{XLM1L\_AttAvgFCMaxFCTanh}   & 51.40 & 46.17 & 92.33 & 61.55 \\
        & \texttt{XLM1L\_AttAvgFCMaxFCSum}    & 53.55 & 47.24 & 87.62 & 61.39 \\
        & \texttt{XLM1L\_AttAvgFCMaxFC}       & 48.37 & 44.79 & \textbf{96.73} & 61.22 \\
        & \texttt{XLM1L\_AttAvgFC}            & 52.31 & 46.63 & 91.05 & 61.67 \\
        & \texttt{XLM1L\_AttMaxFC}            & \textbf{54.90} & \textbf{48.06} & 87.06 & 61.93 \\
        & \texttt{XLM1L\_Att}                 & 52.84 & 46.89 & 89.78 & 61.61 \\
        
        \midrule
        
        \multirow{8}{*}{\shortstack[l]{EN-S}} 
        & \wip{}                             & 71.27 & 61.65 & \textbf{84.14} & 71.16 \\
        & \texttt{XLM1L\_AttAvgFCMaxFCVarFC}  & 73.28 & 64.79 & 80.05 & \textbf{71.62} \\
        & \texttt{XLM1L\_AttAvgFCMaxFCTanh}   & 74.05 & 67.72 & 73.34 & 70.42 \\
        & \texttt{XLM1L\_AttAvgFCMaxFCSum}    & 73.00 & 64.38 & 80.29 & 71.46 \\
        & \texttt{XLM1L\_AttAvgFCMaxFC}       & 74.02 & 68.19 & 71.81 & 69.96 \\
        & \texttt{XLM1L\_AttAvgFC}            & \textbf{75.05} & \textbf{69.13} & 73.64 & 71.31 \\
        & \texttt{XLM1L\_AttMaxFC}            & 73.79 & 67.28 & 73.52 & 70.26 \\
        & \texttt{XLM1L\_Att}                 & 70.76 & 63.40 & 72.30 & 67.56 \\
        \midrule
        
        \multirow{8}{*}{ES}
        & \wip{}            & 68.81 & 58.16 &\textbf{ 86.97 }& \textbf{69.70} \\
        & \texttt{XLM1L\_AttAvgFCMaxFCVarFC}  & 69.62 &  59.55 & 64.22 & 61.79 \\
        & \texttt{XLM1L\_AttAvgFCMaxFCTanh}   & 69.25 & 60.69 & 72.27 & 65.98 \\
        & \texttt{XLM1L\_AttAvgFCMaxFCSum}    & 64.63 & 54.63 & 84.09 & 66.23 \\
        & \texttt{XLM1L\_AttAvgFCMaxFC}       & \textbf{73.01} & \textbf{65.07} & 68.15 & 67.62 \\
        & \texttt{XLM1L\_AttAvgFC}            & 70.44 & 62.97 & 68.79 & 65.75 \\
        & \texttt{XLM1L\_AttMaxFC}            & 69.81 & 64.39 & 60.00 & 62.12 \\
        & \texttt{XLM1L\_Att}                 & 68.12 & 60.81 & 63.94 & 62.33 \\
        
        \bottomrule
    \end{tabular}
    \label{tab:xlm_AXEL_ablation}
\end{table*}

\end{document}